\pdfoutput=1

\documentclass[11pt]{article}

\usepackage[]{acl2023}

\usepackage{times}
\usepackage{latexsym}

\usepackage[T1]{fontenc}
\usepackage[utf8]{inputenc}

\usepackage{microtype}

\usepackage{inconsolata}

\usepackage{xspace}
\usepackage{graphicx}
\usepackage{multirow}
\usepackage{multicol}
\usepackage{booktabs}
\usepackage{amsmath}
\usepackage{makecell}
\usepackage[inkscapelatex=false]{svg}

\newcommand{\abr}[1]{\textsc{#1}}
\newcommand\ours{\abr{DTE}\xspace}
\newcommand\oursdataset{\textsc{NoisySP}\xspace}

\newcommand\Unk{Unanswerable\xspace}

\newcommand\AMBadj{Ambiguous\xspace}

\newcommand{\reftab}[1]{Table~\ref{#1}}
\newcommand{\reffig}[1]{Figure~\ref{#1}}
\newcommand{\refsec}[1]{Sec.~\,\ref{#1}}
\newcommand{\refapp}[1]{Appendix\,\ref{#1}}

\newcommand{\question}{Q}
\newcommand{\word}{q}
\newcommand{\sColumnSet}{\mathcal{C}}
\newcommand{\sColumn}{c}

\newcommand{\tagg}{L}
\newcommand{\atag}{l}

\newcommand*{\affaddr}[1]{#1}
\newcommand*{\affmark}[1][*]{\textsuperscript{#1}}
\renewcommand{\tt}[1]{\fontfamily{cmtt}\selectfont #1}

\title{\textit{Know What I don't Know}: \\ Handling Ambiguous and Unanswerable Questions for Text-to-SQL}

\author{
\makecell{Bing Wang\affmark[\textdagger]{\thanks{Work done during an internship at Microsoft Research Asia.}}~~, Yan Gao\affmark[\S], Zhoujun Li\affmark[\textdagger], Jian-Guang Lou\affmark[\S]}
\\
\centerline{\affaddr{\affmark[\textdagger]State Key Lab of Software Development Environment, Beihang University, Beijing, China}} \\
\centerline{\affaddr{\affmark[\S]Microsoft Research Asia}}
\\
\centerline{
    \tt {\{bingwang, lizj\}@buaa.edu.cn}, \tt {\{yan.gao, jlou\}@microsoft.com}
}}

\begin{document}
\maketitle
\begin{abstract}

The task of text-to-SQL aims to convert a natural language question into its corresponding SQL query within the context of relational tables. Existing text-to-SQL parsers generate a ``plausible'' SQL query for an arbitrary user question, thereby failing to correctly handle problematic user questions. To formalize this problem, we conduct a preliminary study on the observed ambiguous and unanswerable cases in text-to-SQL and summarize them into 6 feature categories.  Correspondingly, we identify the causes behind each category and propose requirements for handling ambiguous and unanswerable questions. Following this study, we propose a simple yet effective counterfactual example generation approach that automatically produces ambiguous and unanswerable text-to-SQL examples. Furthermore, we propose a weakly supervised \textsc{DTE} (\textbf{D}etecting-\textbf{T}hen-\textbf{E}xplaining) model for error detection, localization,  and explanation. Experimental results show that our model achieves the best result on both real-world examples and generated examples compared with various baselines. We release our data and code at: \href{https://github.com/wbbeyourself/DTE}{https://github.com/wbbeyourself/DTE}.

\end{abstract}

\section{Introduction}\label{sec:intro}

Text-to-SQL task aims to generate an executable SQL query given a natural language (NL) question and corresponding tables as inputs.
It builds a natural language interface to the database to help users access information in the database ~\cite{Popescu2003TowardsAT}, thereby receiving considerable interest from both industry and academia~\cite{guo-etal-2019-towards, wang-etal-2020-rat, liu-etal-2021-awakening}.
Correspondingly, a series of new model architectures have been proposed, such as IRNet~\cite{guo-etal-2019-towards}, RAT-SQL~\cite{wang-etal-2020-rat}, ETA~\cite{liu-etal-2021-awakening}, etc. 
These models have achieved satisfactory results on well-known benchmarks, including Spider~\cite{yu-etal-2018-spider} and WikiSQL~\cite{zhong2017seq2sql}.

\begin{figure}[t]
        \centering
      \includegraphics[width=1.0\linewidth]{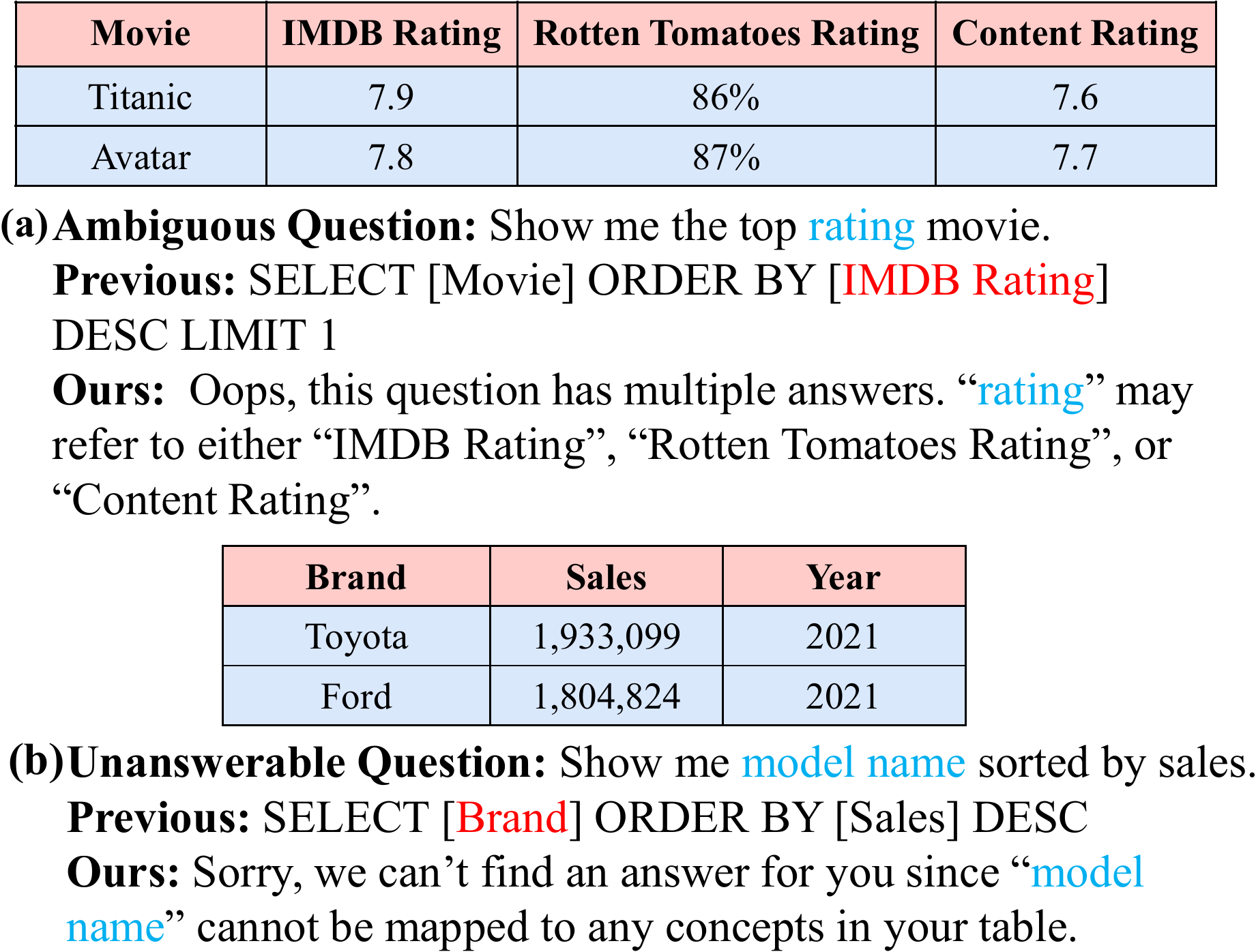}
       \caption{\AMBadj and unanswerable examples in text-to-SQL task as well as our explanations. Blue font denotes the problematic question span and red font means the ``plausible'' column name selected by previous models.}
        \label{fig:intro-example}
\end{figure}

However, state-of-the-art models trained on the leaderboard datasets still demonstrate inadequate performance in practical situations, where user queries are phrased differently, which can be problematic.
Concretely, from our study with real-world text-to-SQL examples (\refsec{sec:study}), it is found that about 20\% of user questions are problematic, including but not limited to \textit{ambiguous} and \textit{unanswerable} questions. 
Ambiguous questions refer to those which can have multiple semantic meanings based on a single table. 
For instance, in \reffig{fig:intro-example}(a), the word ``rating'' in a user's query could be mapped to disparate columns, such as ``IMDB Rating'', ``Rotten Tomatoes Rating'', or ``Content Rating''. 
On the other hand, unanswerable questions pertain to those that cannot be answered based on the information provided by the tables.
For example, in \reffig{fig:intro-example}(b), there is no column about ``model name'' in the table.  
State-of-the-art models are capable of generating ``plausible'' SQL queries, even in the presence of ambiguous or unanswerable questions.

This phenomenon reveals two problems of previous methods. 
Firstly, with regard to data, the training samples utilized in these approaches lack ambiguous and unanswerable questions.
Current training datasets gather queries by either using templates~\cite{zhong2017seq2sql} or by manually annotating controlled questions and filtering out poorly phrased and ambiguous ones~\cite{yu-etal-2018-spider}.
This data-gathering approach ensures that a correct answer exists within the table context.
Secondly, in regards to the model, end-to-end parsing models ignore modeling questions in a fine-grained manner, which results in an inability to precisely detect and locate the specific reasons for ambiguous or unanswerable questions.

To address the data shortage problem, we propose a counterfactual examples generation approach that automatically produces ambiguous and unanswerable text-to-SQL examples using existing datasets.
Given the free-form nature of the text, conventional natural language modification techniques are not always accurate.
In contrast to plain text, tables exhibit well-defined structures, usually consisting of rows and columns.
Consequently, table modification is more controllable.
In light of this, we propose to generate ambiguous and unanswerable examples by modifying the structured table.

Furthermore, we propose a weakly supervised model \ours (\textbf{D}etecting-\textbf{T}hen-\textbf{E}xplaining) for handling ambiguous and unanswerable questions.
To locate ambiguous or unanswerable tokens in user questions, we formulate the location process as a sequence labeling problem, where each token in the user question will be tagged as being related to an ambiguous label, an unanswerable label, or others (\refsec{sec:task-definition}).
Since there is no labeled data for sequence labeling, we extract the set of column names and cells appearing in the SQL query and use this set as the weak supervision.
In this way, we could generate explicit explanations for ambiguous and unanswerable questions to end users.
Note that the sequence labeling information is pseudo and derived from our model, thus alleviating  heavy manual efforts for annotation.

\begin{table*}[htbp]
\small
    \centering
    \scalebox{0.8}{
        \begin{tabular}{c}
            \includegraphics[width=1.0\textwidth]{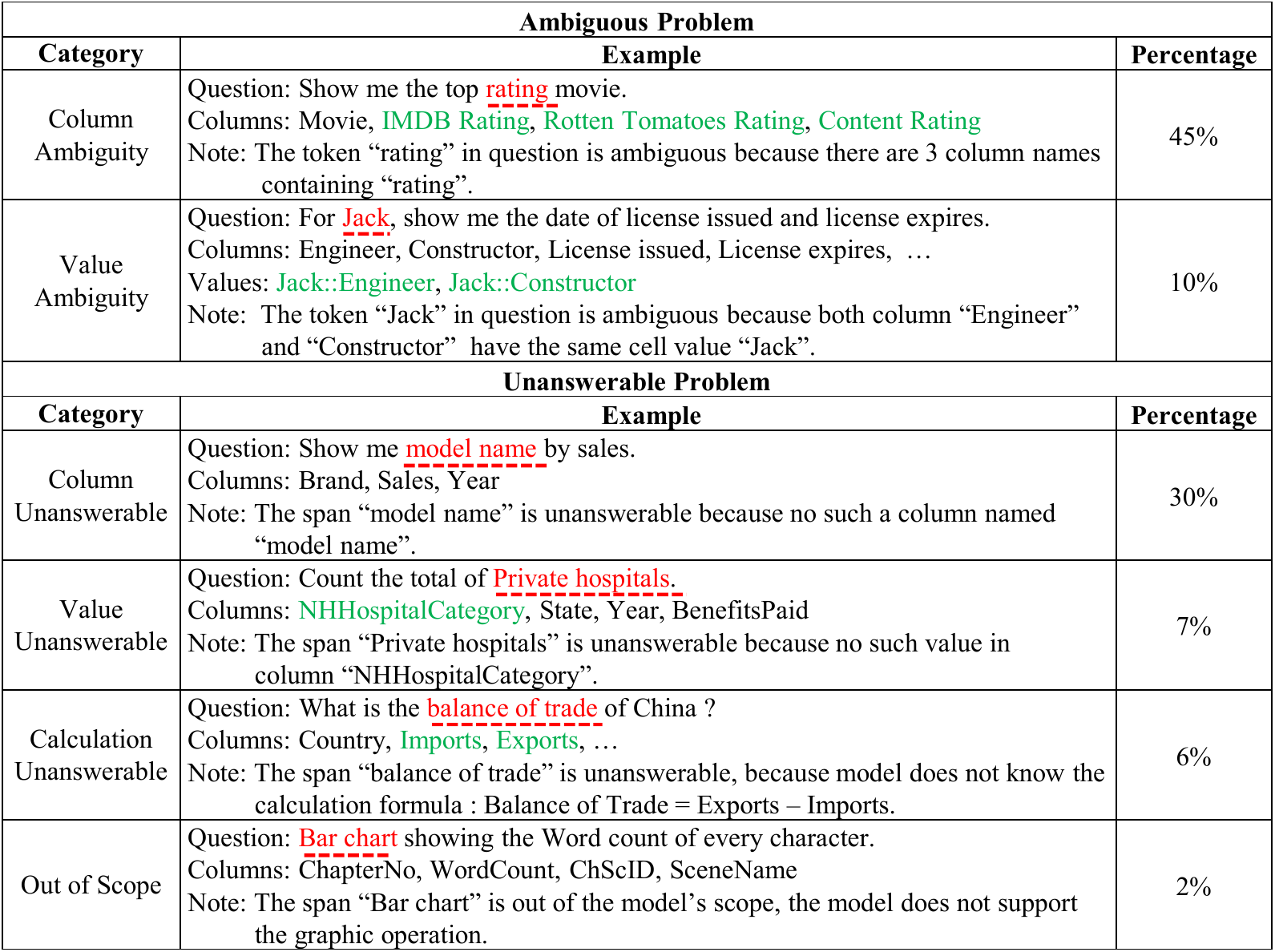}
        \end{tabular}
    }
	\caption{\AMBadj and unanswerable problem categories in text-to-SQL task. The red font with a dashed line denotes the ambiguous or unanswerable question span. The green font means a related concept (columns or values) to the red span. Note sentence explains why the example is ambiguous or unanswerable. }\label{table:category}
\end{table*}

Experimental results show that our approach achieves the best results on both real-world examples collected from realistic applications and automatically generated ambiguous and unanswerable examples, compared with various baselines.
Our contributions are as follows:

\begin{itemize}
    \item We conduct a preliminary study on the ambiguous and unanswerable questions in text-to-SQL and summarize 6 featured categories. We also identify the causes behind each category and propose requirements that should be met in explainable text-to-SQL systems.

    \item We propose a counterfactual examples generation approach that automatically produces ambiguous and unanswerable text-to-SQL examples via modifying structured tables.
    
    \item We propose a weakly supervised model for ambiguous and unanswerable question detection and explanation. 
    Experimental results show that our approach brings the model with the best explainability gain compared with various baselines. 
\end{itemize}

\section{Preliminary Study on Ambiguous and Unanswerable Problem}\label{sec:study}

To understand user behaviors in a real-world application, we conduct a comprehensive user study on our commercial text-to-SQL product.
Firstly, around 3,000 failed user questions in the product are collected.
They obtained over 30 data tables from multiple domains, including education, finance, government, etc.
Then, we manually group these questions into multiple categories.
At last, we explore the causes and potential solutions to deal with them.
According to our analysis, nearly 20\% of the questions are problematic, including 55\% ambiguous and 45\% unanswerable questions respectively, revealing the importance of handling problematic questions. 
In the following, we will introduce their categories, causes, and potential solutions for handling them.

\subsection{Problem Categories}
\label{sec:problem-definition}
In this section, we formalize ambiguous and unanswerable questions and identify 6 sub-categories.

\paragraph{\AMBadj Problem} 
In the text-to-SQL task, ambiguity means that one user question could have multiple semantic meanings (e.g., SQL query) based on one table.
Specifically, we can subdivide them into two sub-categories, namely column ambiguity and value ambiguity, which account for 45\% and 10\% of all problematic questions, respectively.
Column ambiguity means that some tokens in the user question could be mapped to multiple columns.
For example in \reftab{table:category}, we don't know exactly which ``Rating'' the user wants since there are three \textit{rating} columns.
Value ambiguity means that some tokens in the user question could be mapped to multiple cell values in the table.
For example in \reftab{table:category}, \textit{Jack} in the user question can be mapped to the name of either an ``Engineer'' or a ``Constructor''.

\paragraph{\Unk Problem} 
The unanswerable problem can be classified into four categories: column unanswerable, value unanswerable, calculation unanswerable, and out-of-scope, which account for 30\%, 7\%, 6\%, and 2\% of all problematic questions, respectively, as shown in the bottom part of \reftab{table:category}.
(1) The column unanswerable means that the concepts mentioned in the question do not exist in table columns. 
In the first example, the \textit{model name} does not exist in the given columns, but our product incorrectly associates it with the irrelevant column ``Brand''.
(2) The value unanswerable indicates that the user question refers to cell values that do not exist in the table.
As the second example shows, no such \textit{Private hospitals} value exists in the table.
(3) The calculation unanswerable category is more subtle. 
It requires mapping the concept mentioned in the user question to composite operations over existing table columns.
For example, the \textit{balance of trade} is a concept derived from ``$Exports - Imports$''.
Such mapping functions require external domain knowledge. 
Our product which is trained from a general corpus captures limited domain knowledge, and thus often fails.
(4) The out-of-scope category means that the question is out of SQL's operation scope, such as chart operations.

\subsection{Causes}
\label{sec:causes}
Through communicating with end users and analyzing the characteristic of questions as well as corresponding table contexts, we identify three fundamental causes for ambiguous and unanswerable questions:
(1) end users are unfamiliar with the content of the table and don't read the table carefully, causing unanswerable questions;
(2) ambiguity arises due to the richness of natural language expressions and the habitual omission of expressions by users~\citep{radhakrishnan-etal-2020-colloql}; (3) the emergence of similar concepts in  the table tends to cause more ambiguous questions.
Note that around 95\% of problematic questions are constructed unintentionally, revealing the importance of making users \textit{conscious of being wrong}.

\begin{figure}[t]
\small
    \centering
      \includegraphics[width=1.0\linewidth]{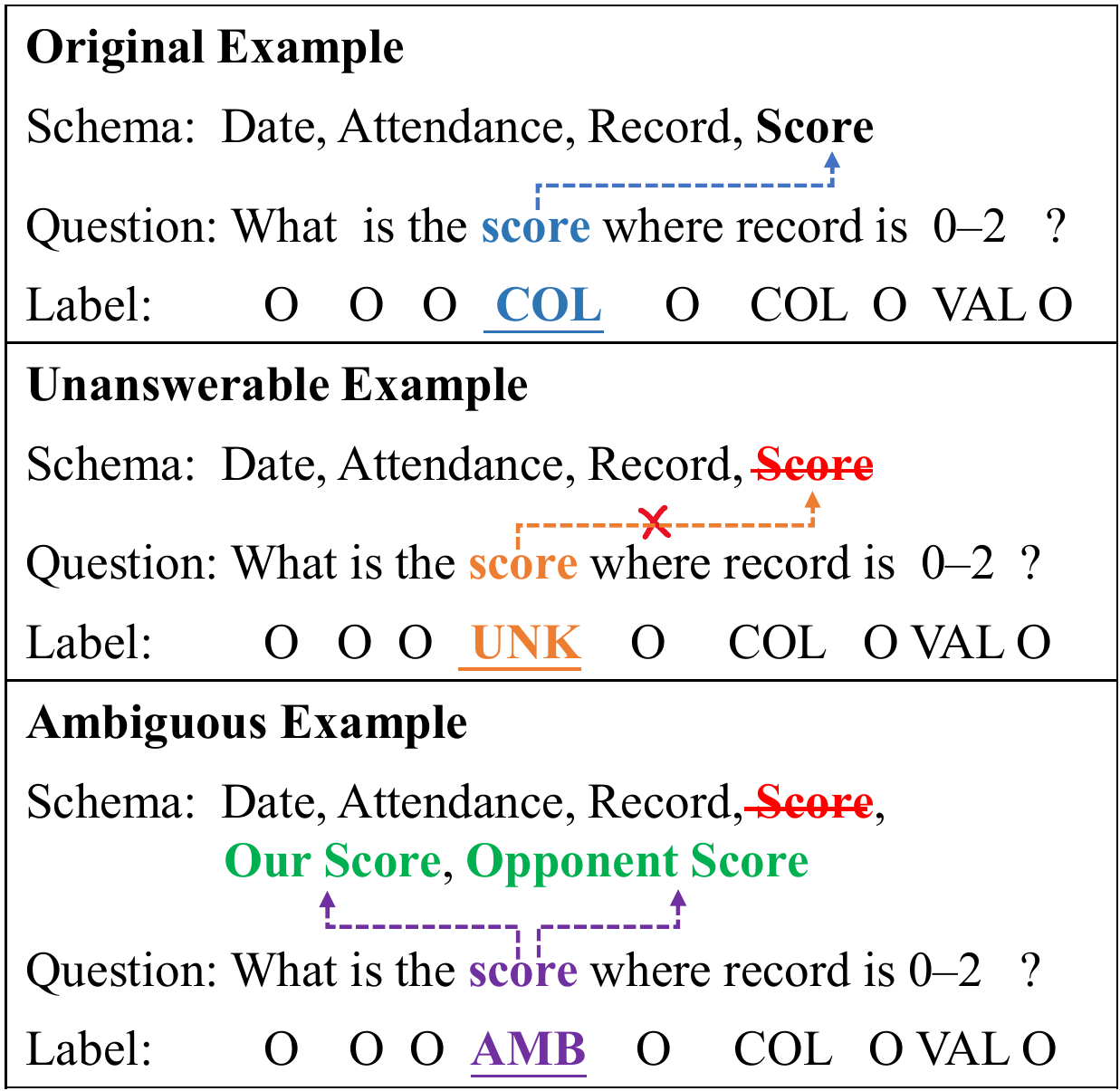}
       \caption{\AMBadj and unanswerable examples generated by our approach.}
        \label{fig:aug-example}
\end{figure}

\subsection{Explainable Parser Requirements}

Based on the findings and analysis above, to deal with ambiguous and unanswerable questions, we propose to make a text-to-SQL system \textit{know-what-I-don't-know}. 
On one hand, a parsing system should detect ambiguous and unanswerable questions. On the other hand, a parsing system should locate the specific reasons and generate corresponding explanations to guide the user in rectification.

Achieving \textit{know-what-I-don't-know} can benefit from two aspects: (1) from model view: enhances models' ability to deal with problematic questions and improve user trust; (2) from user view: makes it clear to users which part of their questions are problematic, guiding them to revise their questions.
In our user study experiments, we find that 90\% of the problematic questions can be corrected by prompting users with explanations shown in \reftab{table:category}, and the remaining 10\% of  questions can only be solved by injecting external knowledge into the model.
In the following, we will introduce how we mitigate the challenges mentioned in \refsec{sec:intro}

\section{Counterfactual Examples Generation}\label{sec:data-generation}

To alleviate the data shortage issue, we propose a counterfactual examples generation approach for automatically generating problematic text-to-SQL examples.
In our approach, we mainly focus on generating two major types of problematic examples: column ambiguity and column unanswerable, which account for 75\%~\footnote{Proposal for handling the remaining 25\% questions can be found in Appendix} of all problematic examples based on our preliminary study.
Note that the counterfactual examples are generated via modifying structured tables instead of natural language questions.
The reason is that conditional modification on a structured table is more controllable than unstructured text.
Finally, 23k problematic examples are obtained based on two text-to-SQL datasets, i.e., WikiSQL~\citep{zhong2017seq2sql} and WTQ~\citep{shi-etal-2020-potential}.
Next, we will introduce the details of our approach.

\subsection{Our Approach}

Given an answerable text-to-SQL example that contains a question $\question = ( \word_1, \dots, \word_m )$, a DB schema (also a column set) $\sColumnSet = \{ \sColumn_1, \dots, \sColumn_{n} \}$ and a SQL query $S$, our goal is to generate problematic examples, denoting as ($Q$, $\sColumnSet^{'}$, $S$) triplets.
By removing evidence supporting $Q$ from $\sColumnSet$ or adding ambiguous ones, a new DB schema $\sColumnSet^{'}$ is generated.

\paragraph{\Unk Examples Generation} Specifically, we randomly sample a target column $\sColumn_t$ in the SQL query $S$.
Then we delete $\sColumn_t$ from $\sColumnSet$ to remove the supporting evidence for question spans $Q_{s} = ( \word_i, \dots, \word_j )$ that mentioned $\sColumn_t$.
At last, the question span $Q_{s}$ is labeled as \texttt{UNK}. 
For instance, in the unanswerable example of \reffig{fig:aug-example}, given an original  question ``What is the score where record is 0–2?'', the question span ``score'' is grounded to the column ``Score''.
By deleting the column ``Score'', we obtain an unanswerable example.

\begin{table}[t]
\small
  \centering
    \begin{tabular}{l|c|c|c}
    \hline
                    & NoisySP & WikiSQL & WTQ \\
    \hline
    \textbf{Train}  &        &       &  \\
    \# ambiguous      & 4,760  & 0     & 0     \\
    \# unanswerable         & 10,673 & 0     & 0  \\
    \# answerable   &   0    & 56,350 & 7,696 \\
    \# tables       & 4,861  & 17,984  & 1,283  \\
    \hline
    \textbf{Development} &       &       &  \\
    \# ambiguous           & 1,581 & 0     & 0     \\
    \# unanswerable              & 1,652 & 0     & 0  \\
    \# answerable        & 0     & 8,142 & 1,772  \\
    \# tables            & 1,232 & 2,614 & 325  \\
    \hline
    \textbf{Test} &        &       &  \\
    \# ambiguous    & 2,332 & 0     & 0    \\
    \# unanswerable       & 2,560  & 0     & 0  \\
    \# answerable & 0      & 15,362 & 0   \\
    \# tables     & 1,993  & 5,031  & 0  \\
    \hline
    \end{tabular}
\caption{Dataset statistics of \oursdataset, compared to the original WikiSQL and WTQ dataset.}
  \label{tab:dataset-statistics}
\end{table}

\paragraph{\AMBadj Examples Generation} Similar to unanswerable examples generation, we generate an ambiguous example by firstly deleting a column $\sColumn_t$ and then adding two new columns. 
The critical point is that newly added columns are expected to (1) fit nicely into the table context; (2) have high semantic associations with the target column $\sColumn_t$ yet low semantic equivalency (e.g. ``opponent score'' is semantically associated with ``score'', but it is not semantic equivalent).
To achieve this, we leverage an existing contextualized table augmentation framework, CTA~\citep{pi-etal-2022-towards}, tailored for better contextualization of tabular data, to collect new column candidates.
We select target columns from within the SQL, and typically choose 2-3 near-synonyms for each column candidate.
After that, we rerank the column candidates by their length and similarity with the column $\sColumn_t$, and keep the top 2 as our newly added columns.
As shown in the ambiguous example of \reffig{fig:aug-example}, we first  delete the  original  column ``Score'', then add two domain-relevant and semantically associated columns  ``Our Score'' and ``Opponent Score''.

\subsection{Dataset Statistic}

Leveraging our counterfactual examples generation approach, we obtain a dataset, called \oursdataset based on two cross-domain text-to-SQL datasets, i.e., WikiSQL~\citep{zhong2017seq2sql} and WTQ~\citep{shi-etal-2020-potential}.
Consistent with our preliminary study, we generate 20\% of the original data count as problematic examples.
Finally, we get 23k problematic examples. 
Detailed statistics can be seen in \reftab{tab:dataset-statistics}.
To ensure the quality of the development set and test set, we hired 3 annotators to check the candidate set of newly added columns for ambiguous examples and then drop low-quality ones.
Note that the rate of low quality is only 5\%, demonstrating the effectiveness of our approach. 

\begin{figure*}[t]
        \centering
        \includegraphics[width = 0.9\linewidth]{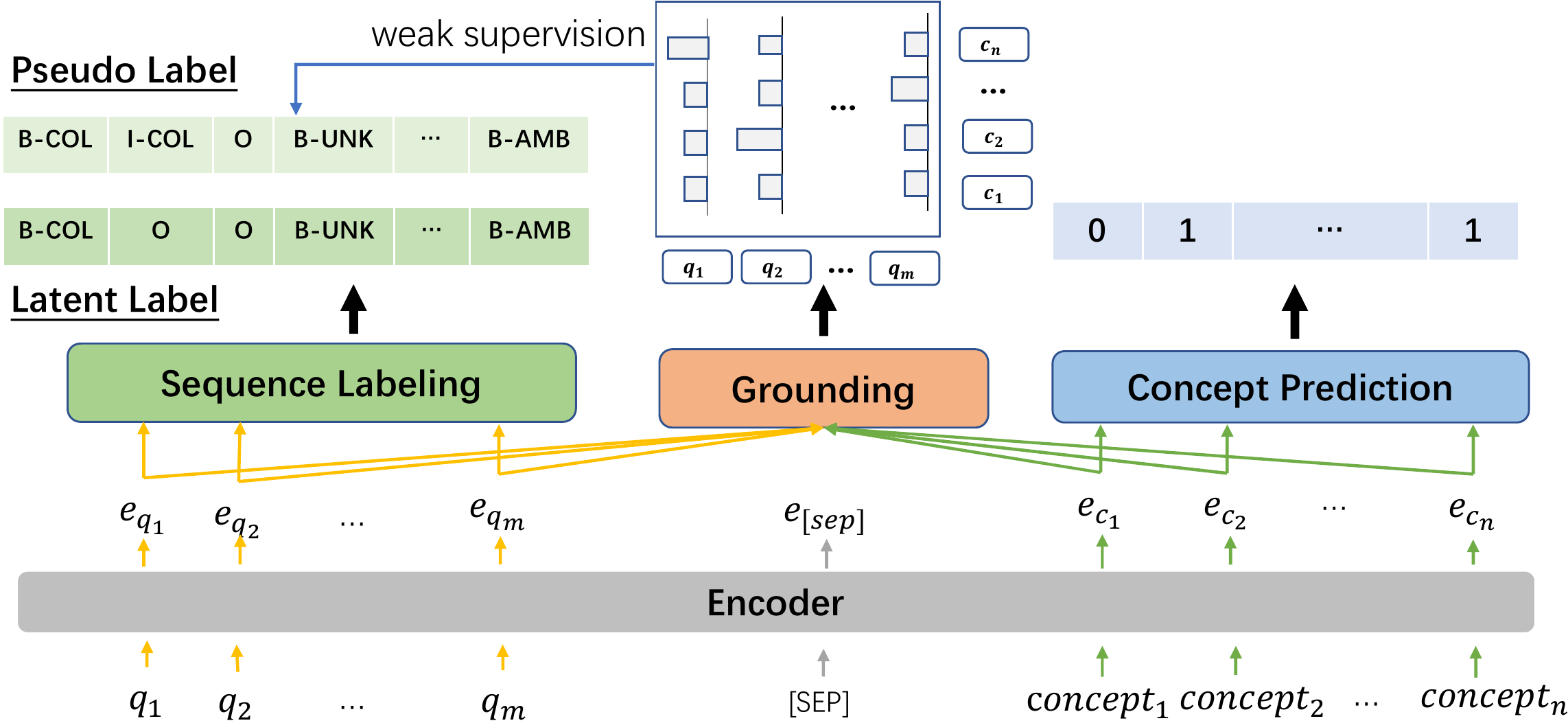}
        \caption{The overall architecture of \ours.}
        \label{fig:model}
\end{figure*}

\begin{table}[t]
\small
  \centering
  \scalebox{1.0}{
    \begin{tabular}{lll}
    \toprule
    \textbf{Labels} & \textbf{Description} & \textbf{Example(Token:Label)} \\
    \midrule
    COL  & Column Mention  & sales: B-COL \\
    VAL  & Value Mention   & godfather: B-VAL \\
    AMB  & \AMBadj Span  & rating: B-AMB \\
    UNK  & \Unk Span    & model: B-UNK \\
    O    & Nothing           & the: O \\
    \bottomrule
    \end{tabular}
    }
\caption{Labeling categories of question tokens }
 \label{tab:labels}
\end{table}

\section{Model: \ours}\label{sec:method}

In this section, we introduce our \textbf{D}etecting-\textbf{T}hen-\textbf{E}xplaining (DTE) model to handle ambiguous and unanswerable questions.
To generate a fine-grained explanation, we formulate it as a sequence labeling problem, where each token in the user question will be tagged as being related to an ambiguous label, an unanswerable label, or others.
Concretely, \ours consists of three modules: concept prediction module, grounding module, and sequence labeling module. 
The grounding module generates pseudo-label information to guide the training of the sequence labeling module.
The overall architecture of \ours is shown in \reffig{fig:model}.

\subsection{Task definition}\label{sec:task-definition}

Given an input question $\question = ( \word_1, \dots, \word_m )$, a data table (with a concept set $C = {c_1, \dots, c_k}$, containing columns and cell values),
the goal of sequence labeling is to output a labeling sequence $\tagg = (\atag_1, \dots, \atag_m$) for each token in $\question$.
It can be represented by tagging each token in the question with a set of BIO labels~\citep{tjong-kim-sang-veenstra-1999-representing}. 
Specifically, we define 5 kinds of labels for question tokens, namely COL, VAL, AMB, UNK, and O. 
Their descriptions and examples are shown in \reftab{tab:labels}.

\subsection{Preliminaries: ETA for grounding}\label{sec:preliminary}

In this work, we formulate problematic question detection as a sequence labeling task, whose training process requires large-scale label annotations as supervision.
However, such annotations are expensive and time-consuming.
To obtain label information in an efficient and cheap way, we propose to leverage the grounding result of the text-to-SQL task and transform it into a pseudo-labeling sequence.
Particularly, we use ETA~\citep{liu-etal-2021-awakening}, a pretrained probing-based grounding model, as the backbone of our approach. 
The major advantage of ETA is that, compared with models relying on expensive annotations of grounding, it only needs supervision that can be easily derived from SQL queries.

\subsection{Sequence Labeling Module}

To meet the requirements of detecting and locating ambiguous and unanswerable question spans, we design a sequence labeling module, which is intuitively suitable for our sequential modeling purpose.
The sequence labeling module consists of a dropout layer, a linear layer, and a CRF layer, following best practices  in previous work ~\citep{yang-etal-2018-design}.
Given a contextualized embedding sequence $(e_{q_1}, \dots, e_{q_m})$, the goal of the sequence labeling module is to output the label sequence $\tagg = \atag_1, \dots, \atag_m$ with the highest likelihood probability.

\subsection{Multi-Task Training}

Our multi-task training process involves three steps: 
(1) train the concept prediction module. 
(2) warm-up grounding module to get alignment pairs. 
(3) train the sequence labeling module with the pseudo tag derived from the grounding module.

\subsection{Response Generation}
At the inference step, given a question and the table information, \ours predicts labels for each question token and outputs grounding pairs between question tokens and table entities.
If the \texttt{AMB} (or \texttt{UNK}) label occurs, it means it is an ambiguous (or unanswerable) question.
To generate corresponding interpretations to end users, we carefully design two response templates.
More details about the templates could be found in \refapp{sec:respnd-templ}.

\section{Experiments}\label{sec:exps}

In this section, we systematically evaluate the effectiveness of \ours. 
Specifically, we examine \ours's performance in two aspects: (1) the performance of the sequence labeling module in detecting ambiguous and unanswerable tokens;  (2) the grounding performance for each label to provide evidence for generating explainable responses to end users.
In addition, we report the evaluation results on text-to-SQL tasks.

\subsection{Experimental Setup}\label{sec:exp-setup}

\paragraph{Datasets}
We conduct experiments based on the following datasets: 
(1) \oursdataset with 23k automatically generated examples, 
(2) two cross-domain text-to-SQL datasets, i.e.,  WikiSQL~\citep{zhong2017seq2sql} and WTQ~\citep{shi-etal-2020-potential}\footnote{Note that we use the version with SQL annotations provided by \citet{shi-etal-2020-potential} , since the original WTQ~\citep{pasupat-liang-2015-compositional} only contains answer annotations.}, 
(3) 3,000 real-world examples collected by us (\refsec{sec:study}).
All models are trained with the \oursdataset, WikiSQL, and WTQ datasets.
Specifically, real-world examples are only used for testing. 
Dataset statistics are shown in \reftab{tab:dataset-statistics}.

\paragraph{Evaluation Metric}
To evaluate sequence labeling performance, we report accuracy for each label category.
For grounding performance evaluation, we report grounding accuracy for each label, except for UNK and O, which have no grounding results.

\paragraph{Baseline Models}

We choose two types of representative models for comparison: (1) the heuristic-based method~\citep{sorokin-gurevych-2018-mixing}, which is widely used in entity linking and grounding tasks; (2) the learning-based method, ETA~\citep{liu-etal-2021-awakening}, which is a strong grounding baseline, leveraging the intrinsic language understanding ability of pretrained language models. 
We update them with a little modification to fit our task because their vanilla version is not directly applicable.
More implementation details about the baseline and \ours could be found in \refapp{sec:model-impl}.

\subsection{Experimental Results on \oursdataset}\label{sec:exp-results}

\paragraph{Sequence Labeling Results}
As shown in \reftab{tab:label-accuracy}, we compare the performances of \ours with various baselines on the test set of \oursdataset.
\ours outperforms previous baselines across all label categories, which demonstrates the superiority of our \ours model.
Compared with the heuristic-based method, \ours significantly improves performances by 25\% average gains of all label categories, which shows that our \oursdataset dataset is challenging and the heuristic-based method is far from solving these questions.
Besides, \ours consistently outperforms the ETA+BERT baseline by a large margin, not only improving the ambiguous and unanswerable label accuracy by 7\% and 11\%, respectively, but also improving column and value detecting accuracy, demonstrating the effectiveness of our approach for detection.

\begin{table}[t]
\small
  \centering
    \begin{tabular}{lccccc}
    \toprule
    Models & COL   & VAL   & AMB   & UNK   & O \\
    \midrule
    Heuristic & 61.7  & 66.8  & 57.8  & 60.7  & 72.1 \\
    ETA+BERT & 83.4  & 87.9  & 75.6  & 70.2  & 80.9 \\
    ETA+BERT$_L$ & 85.7  & 90.4  & 76.4  & 71.4  & 82.7 \\
    \midrule
    \ours+BERT & 88.2  & 94.1  & 81.4  & 78.6  & 90.7 \\
    \ours+BERT$_L$ & \textbf{89.4} & \textbf{95.7} & \textbf{83.2} & \textbf{80.3} & \textbf{92.4} \\
    \bottomrule
    \end{tabular}
    
\caption{Sequence labeling accuracy of \ours compared with baselines in \oursdataset test set.}
  \label{tab:label-accuracy}
\end{table}

\begin{table}[t]
\small
  \centering
    \begin{tabular}{lccc}
    \toprule
    Models & COL & VAL & AMB \\
    \midrule
    Heuristic    & 55.9  & 67.2  & 56.2 \\
    ETA+BERT & 71.4  & 75.3  & 60.7 \\
    ETA+BERT$_L$ & 72.4  & 77.8  & 62.4 \\
    \midrule
    \ours+BERT & 73.4  & 78.2  & 79.8 \\
    \ours+BERT$_L$ & \textbf{75.1} & \textbf{80.7} & \textbf{82.4} \\
    \bottomrule
    \end{tabular}%
  \caption{Grounding accuracy of \ours compared with baselines in \oursdataset test set.}
  \label{tab:grounding-acc}
\end{table}

\paragraph{Grounding Results}
In order to identify the specific reasons for ambiguous questions, we need to not only identify the target spans, but also establish grounding by finding the linked concept (column or value).
As shown in \reftab{tab:grounding-acc}, we compare \ours's grounding performance with various baselines.
Note that the unanswerable span in question does not require grounding to any concept, thus we do not report the grounding result of it.
We observe that \ours consistently outperforms baselines across three label categories on grounding performance, especially on the ambiguous grounding  whose linked concepts are more varied and diverse.

\begin{figure}[t]
\small
        \centering
      \includegraphics[width=1.0\linewidth]{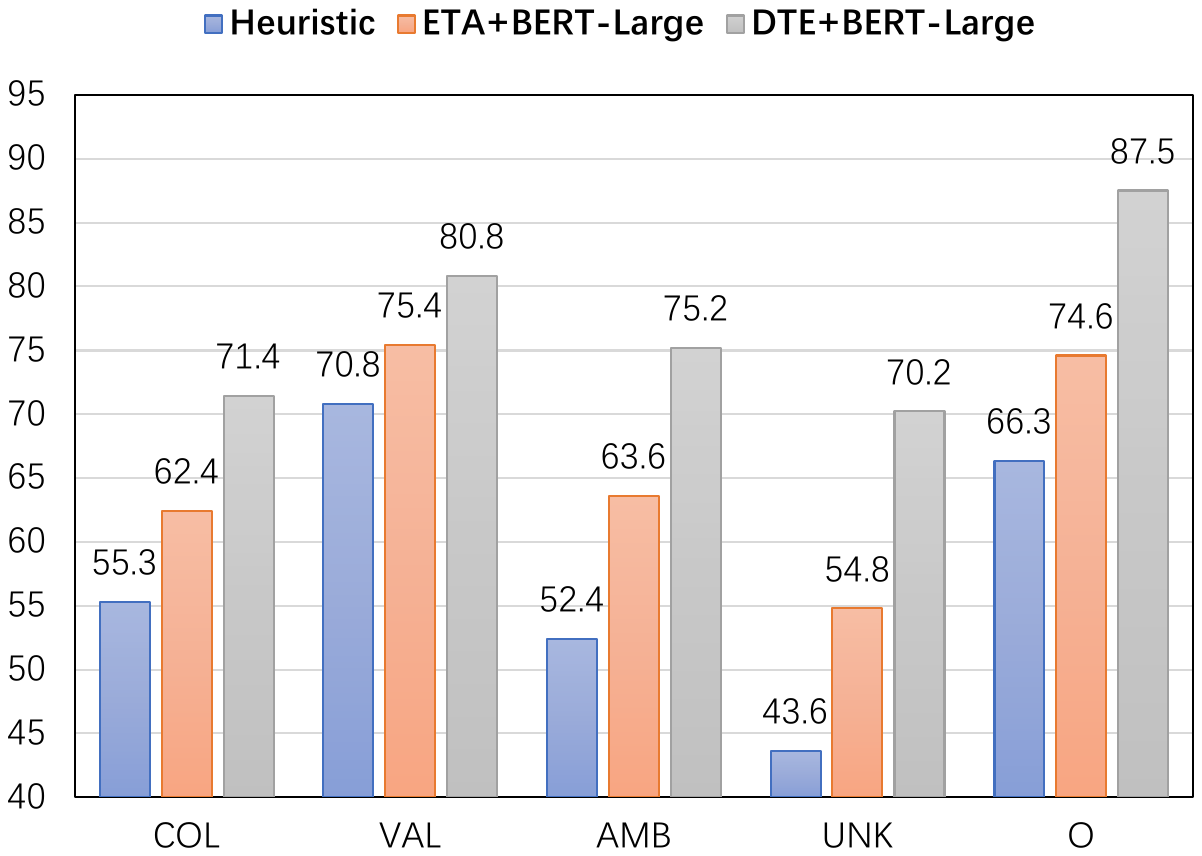}
       \caption{Sequence labeling accuracy of \ours compared with baselines in realistic data.}
        \label{fig:label-accuracy-realistic}
\end{figure}

\begin{figure}[t]
\small
        \centering
      \includegraphics[width=1.0\linewidth]{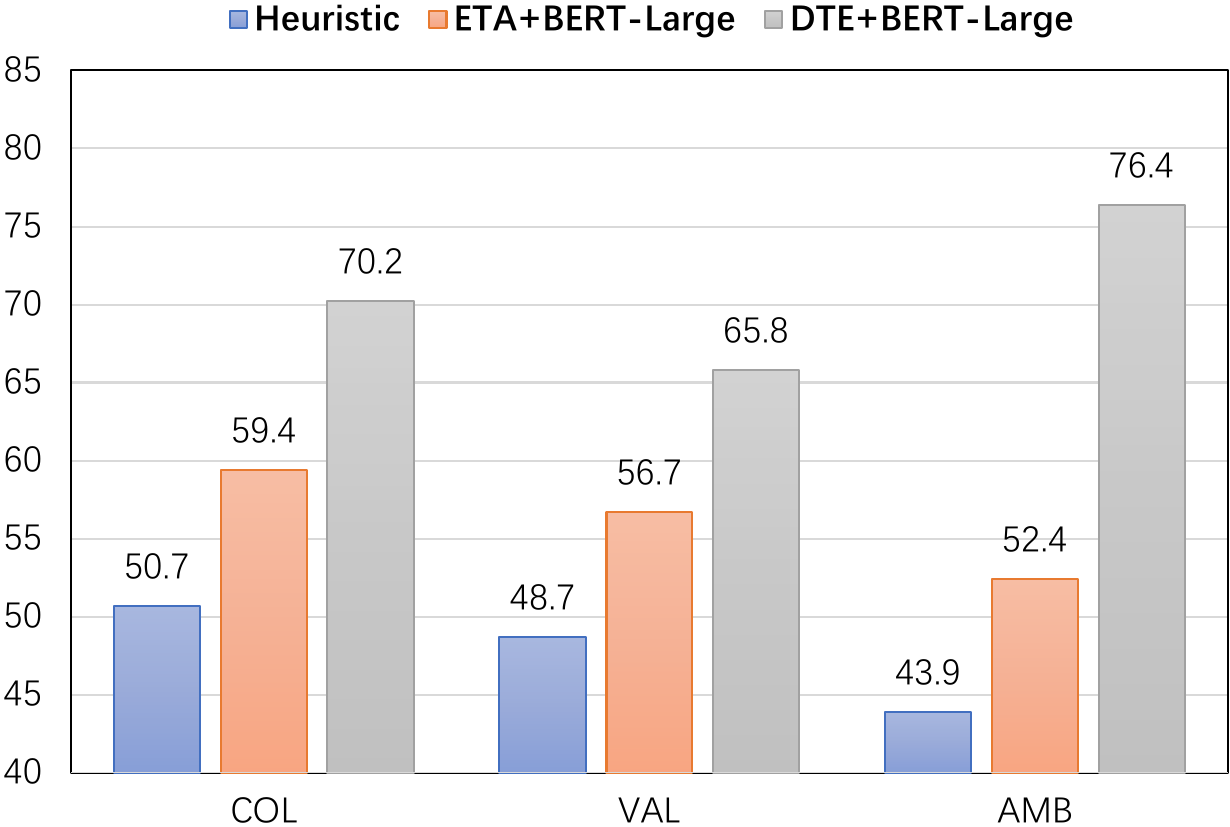}
       \caption{Grounding accuracy of \ours compared with baselines in realistic data.}
        \label{fig:grounding-acc-realistic}
\end{figure}

\subsection{Generalization on Realistic Data}
\label{sec:generaization-to-realistic-data}

We verify the generalization ability of DTE by conducting \textit{out-of-distribution} experiments on 3,000 realistic datasets that we collected from our commercial products. These datasets cover more than 30 data tables from multiple domains such as education, finance, government, etc.
As shown in \reffig{fig:label-accuracy-realistic}, our \ours model still outperforms all baselines consistently and achieves promising performance in ambiguous and unanswerable detection with 75.2\% and 70.2\% sequence labeling accuracy, respectively.
From \reffig{fig:grounding-acc-realistic}, we observe that our \ours model outperforms other baselines by a large margin in grounding accuracy of ambiguous spans.
These results indicate the generalization ability of \ours for handling ambiguous and unanswerable questions and the effectiveness of realistic data.

\begin{table}[htbp]
  \small
    \begin{tabular}{lcccc}
    \toprule
    \multirow{2}[5]{*}{Model} & \multicolumn{2}{c}{Dev} & & Test \\
\cmidrule{2-3} \cmidrule{5-5}          & Ex.Match & Ex.Acc & & Ex.Acc \\
    \midrule
    ALIGN & 37.8  & 56.9       &  & 46.6 \\
    ALIGN+BERT & 44.7  & 63.8  &  & 51.8 \\
    ETA+BERT & 47.6  & \textbf{66.6}    &  & 53.8 \\
    DTE+BERT & \textbf{48.1} & 66.5 & & \textbf{54.2} \\
    \bottomrule
    \end{tabular}%
  \caption{Ex.Match and EX.ACC of text-to-SQL results on the dev and test set of WTQ.}
  \label{tab:text-to-sql}%
\end{table}%

\subsection{Text-to-SQL Results}

To verify the influence of \ours on the text-to-SQL task, we report the exact match accuracy (Ex.Match) and execution accuracy (Ex.Acc) on WTQ dataset.
As shown in \reftab{tab:text-to-sql}, compared with  ALIGN~\citep{lei-etal-2020-examining} and ETA~\citep{liu-etal-2021-awakening}, \ours shows slightly better performance in both exact match accuracy and execution accuracy.
It should be noted that all questions in this experiment are normal questions, i.e., without ambiguous and unanswerable questions.
The result shows that \ours can boost text-to-SQL performance instead of damaging it.

\subsection{Discussion}

\paragraph{What are the remaining errors? }

We manually analyze 20\% of the remaining errors in the \oursdataset dataset and summarize four main error types: (1) wrong detection (25\%) - where our model either misses or over-predicts the ambiguous or unanswerable label. (2) widened span (30\%) - where our model predicts a longer span than the golden result. (3) narrowed span (25\%) - where the model infers a narrowed span than the golden result. (4) other errors (20\%) are caused by the grounding module.
This error analysis indicates the main challenge of \ours is precise localization rather than detection because the second and third errors (55\%) are caused by the wrong span boundary.
More detailed examples can be seen in \refapp{sec:examples-dataset}.

\begin{table}[t]
  \centering
    \begin{tabular}{lccc}
    \toprule
    Models & COL & VAL & O \\
    \midrule
    \ours w/ SL   & \textbf{75.1}  & \textbf{80.7}  & \textbf{90.5} \\
    \ours w/o SL & 72.6  & 75.2  & 82.8 \\
    \bottomrule
    \end{tabular}%
  \caption{Ablation study of \ours model with BERT-large in the grounding accuracy on the test set of \oursdataset. SL means sequence labeling module. }
  \label{tab:ablation-grounding-acc}
\end{table}

\paragraph{Can sequence labeling module benefit grounding module?}
As a multi-task training approach, it is critical to determine the effect of introducing extra tasks on the models' performance on original tasks. 
To verify the influence of the sequence labeling module on grounding results, we conduct an ablation study with or without the sequence labeling module.
As we can see in \reftab{tab:ablation-grounding-acc}, the grounding module does achieve better performance on columns and values alignment with the sequence labeling module.
Through our analysis, we find the performance gain mainly comes from \textit{long concept mention} (with token length > 4).
The reason is that the  CRF layer in the sequence labeling module can strengthen the grounding module's ability to capture long-distance dependency.
In summary, we can conclude that the sequence labeling task can fit in well with the grounding task.

\section{Related Work}
\label{sec:related-work}

\paragraph{Problematic Question Detection.}
Existing works on problematic question detection can be classified into two categories: (1) heuristic-based methods leverage elaborate rules to detect and locate problematic questions span~\citep{sorokin-gurevych-2018-mixing, li-etal-2020-mean, PERQ}, suffering from heavy human efforts on feature engineering.
Besides, some approaches~\cite{dong-etal-2018-confidence, yao-etal-2019-model} estimate the confidence of parsing results, relying on existing parsing models;
(2) on the contrary, learning-based methods don't rely on heuristic rules and parsing models.
For example,  \citet{arthur-etal-2015-semantic} jointly transforms an ambiguous query into both its meaning representation and a less ambiguous NL paraphrase via a semantic parsing framework that uses synchronous context-free grammars.
\citet{zeng-etal-2020-photon} trains a question classifier to detect problematic questions and then employed a span index predictor to locate the position.
However, the index predictor can only find one error span per example, which limits its usage.
In this work, we propose a learning-based approach that could handle multiple errors in problematic questions.

\paragraph{Uncertainty Estimation.}
Recent works on uncertainty estimation of neural networks explore diverse solutions, such as deep ensembles in prediction, calibration, and out-of-domain detection~\citep{liu2020simple}.
However, these methods need network and optimization changes, generally ignore prior data knowledge~\citep{loquercio2020general}, and can only provide uncertainty in predictions without identifying the reasons.
In this work, we propose the counterfactual examples generation approach to adding prior knowledge to the training data and then propose a weakly supervised model for problematic span detection and give explainable reasons.

\section{Conclusion}
\label{sec:Conclusion}

We investigate the ambiguous and unanswerable questions in text-to-SQL and divide them into 6 categories, then we sufficiently study the characteristics and causes of each category.
To alleviate the data shortage issue, we propose a simple yet effective counterfactual example generation approach for automatically generating ambiguous and unanswerable text-to-SQL examples.
What's more, we propose a weakly supervised model for ambiguous and unanswerable question detection and explanation. 
Experimental results verify our model's effectiveness in handling ambiguous and unanswerable questions and demonstrate our model's superiority over baselines.

\section*{Limitations}
Among the six ambiguous and unanswerable problem categories in \reftab{table:category}, our counterfactual example generation approach can not cover the calculation unanswerable and out-of-scope examples generation.
The reason is that our approach focuses on the table transformation ways while generating the calculation unanswerable and out-of-scope examples requires conditional NL modification techniques.
We leave this as our future work.

\section*{Ethics Statement}
Our counterfactual examples generation approach generates a synthesized dataset based on two mainstream text-to-SQL datasets, WikiSQL~\citep{zhong2017seq2sql} and WTQ~\citep{shi-etal-2020-potential}, which are free and open datasets for research use.
All claims in this paper are based on the experimental results.
Every experiment can be conducted on a single Tesla V100.
No demographic or identity characteristics information is used in this paper.

\section*{Acknowledgements}
This work was supported in part by the National Natural Science Foundation of China (Grant Nos. 62276017, U1636211, 61672081), the 2022 Tencent Big Travel Rhino-Bird Special Research Program, and the Fund of the State Key Laboratory of Software Development Environment (Grant No. SKLSDE-2021ZX-18).
We also would like to thank all the anonymous reviewers for their constructive feedback and insightful comments.

% Entries for the entire Anthology, followed by custom entries
\bibliography{anthology,custom}

\begin{thebibliography}{23}
\expandafter\ifx\csname natexlab\endcsname\relax\def\natexlab#1{#1}\fi

\bibitem[{Arthur et~al.(2015)Arthur, Neubig, Sakti, Toda, and
  Nakamura}]{arthur-etal-2015-semantic}
Philip Arthur, Graham Neubig, Sakriani Sakti, Tomoki Toda, and Satoshi
  Nakamura. 2015.
\newblock \href {https://doi.org/10.1162/tacl_a_00159} {Semantic parsing of
  ambiguous input through paraphrasing and verification}.
\newblock \emph{Transactions of the Association for Computational Linguistics},
  3:571--584.

\bibitem[{Dong et~al.(2018)Dong, Quirk, and Lapata}]{dong-etal-2018-confidence}
Li~Dong, Chris Quirk, and Mirella Lapata. 2018.
\newblock \href {https://doi.org/10.18653/v1/P18-1069} {Confidence modeling for
  neural semantic parsing}.
\newblock In \emph{Proceedings of the 56th Annual Meeting of the Association
  for Computational Linguistics (Volume 1: Long Papers)}, pages 743--753,
  Melbourne, Australia. Association for Computational Linguistics.

\bibitem[{Guo et~al.(2019)Guo, Zhan, Gao, Xiao, Lou, Liu, and
  Zhang}]{guo-etal-2019-towards}
Jiaqi Guo, Zecheng Zhan, Yan Gao, Yan Xiao, Jian-Guang Lou, Ting Liu, and
  Dongmei Zhang. 2019.
\newblock \href {https://doi.org/10.18653/v1/P19-1444} {Towards complex
  text-to-{SQL} in cross-domain database with intermediate representation}.
\newblock In \emph{Proceedings of the 57th Annual Meeting of the Association
  for Computational Linguistics}, pages 4524--4535, Florence, Italy.
  Association for Computational Linguistics.

\bibitem[{Lei et~al.(2020)Lei, Wang, Ma, Gan, Lu, Kan, and
  Chua}]{lei-etal-2020-examining}
Wenqiang Lei, Weixin Wang, Zhixin Ma, Tian Gan, Wei Lu, Min-Yen Kan, and
  Tat-Seng Chua. 2020.
\newblock \href {https://doi.org/10.18653/v1/2020.emnlp-main.564} {Re-examining
  the role of schema linking in text-to-{SQL}}.
\newblock In \emph{Proceedings of the 2020 Conference on Empirical Methods in
  Natural Language Processing (EMNLP)}, pages 6943--6954, Online. Association
  for Computational Linguistics.

\bibitem[{Li et~al.(2020)Li, Chen, Liu, Gao, Lou, Zhang, and
  Zhang}]{li-etal-2020-mean}
Yuntao Li, Bei Chen, Qian Liu, Yan Gao, Jian-Guang Lou, Yan Zhang, and Dongmei
  Zhang. 2020.
\newblock \href {https://doi.org/10.18653/v1/2020.emnlp-main.561} {{``}what do
  you mean by that?{''} a parser-independent interactive approach for enhancing
  text-to-{SQL}}.
\newblock In \emph{Proceedings of the 2020 Conference on Empirical Methods in
  Natural Language Processing (EMNLP)}, pages 6913--6922, Online. Association
  for Computational Linguistics.

\bibitem[{Liu et~al.(2020)Liu, Lin, Padhy, Tran, Bedrax~Weiss, and
  Lakshminarayanan}]{liu2020simple}
Jeremiah Liu, Zi~Lin, Shreyas Padhy, Dustin Tran, Tania Bedrax~Weiss, and
  Balaji Lakshminarayanan. 2020.
\newblock Simple and principled uncertainty estimation with deterministic deep
  learning via distance awareness.
\newblock \emph{Advances in Neural Information Processing Systems},
  33:7498--7512.

\bibitem[{Liu et~al.(2021)Liu, Yang, Zhang, Guo, Zhou, and
  Lou}]{liu-etal-2021-awakening}
Qian Liu, Dejian Yang, Jiahui Zhang, Jiaqi Guo, Bin Zhou, and Jian-Guang Lou.
  2021.
\newblock \href {https://doi.org/10.18653/v1/2021.findings-acl.100} {Awakening
  latent grounding from pretrained language models for semantic parsing}.
\newblock In \emph{Findings of the Association for Computational Linguistics:
  ACL-IJCNLP 2021}, pages 1174--1189, Online. Association for Computational
  Linguistics.

\bibitem[{Loquercio et~al.(2020)Loquercio, Segu, and
  Scaramuzza}]{loquercio2020general}
Antonio Loquercio, Mattia Segu, and Davide Scaramuzza. 2020.
\newblock A general framework for uncertainty estimation in deep learning.
\newblock \emph{IEEE Robotics and Automation Letters}, 5(2):3153--3160.

\bibitem[{Pasupat and Liang(2015)}]{pasupat-liang-2015-compositional}
Panupong Pasupat and Percy Liang. 2015.
\newblock \href {https://doi.org/10.3115/v1/P15-1142} {Compositional semantic
  parsing on semi-structured tables}.
\newblock In \emph{Proceedings of the 53rd Annual Meeting of the Association
  for Computational Linguistics and the 7th International Joint Conference on
  Natural Language Processing (Volume 1: Long Papers)}, pages 1470--1480,
  Beijing, China. Association for Computational Linguistics.

\bibitem[{Pi et~al.(2022)Pi, Wang, Gao, Guo, Li, and
  Lou}]{pi-etal-2022-towards}
Xinyu Pi, Bing Wang, Yan Gao, Jiaqi Guo, Zhoujun Li, and Jian-Guang Lou. 2022.
\newblock \href {https://doi.org/10.18653/v1/2022.acl-long.142} {Towards
  robustness of text-to-{SQL} models against natural and realistic adversarial
  table perturbation}.
\newblock In \emph{Proceedings of the 60th Annual Meeting of the Association
  for Computational Linguistics (Volume 1: Long Papers)}, pages 2007--2022,
  Dublin, Ireland. Association for Computational Linguistics.

\bibitem[{Popescu et~al.(2003)Popescu, Etzioni, and
  Kautz}]{Popescu2003TowardsAT}
Ana-Maria Popescu, Oren Etzioni, and Henry~A. Kautz. 2003.
\newblock Towards a theory of natural language interfaces to databases.
\newblock In \emph{IUI '03}, pages 100--112. IEEE.

\bibitem[{Radhakrishnan et~al.(2020)Radhakrishnan, Srikantan, and
  Lin}]{radhakrishnan-etal-2020-colloql}
Karthik Radhakrishnan, Arvind Srikantan, and Xi~Victoria Lin. 2020.
\newblock \href {https://doi.org/10.18653/v1/2020.intexsempar-1.5}
  {{C}ollo{QL}: Robust text-to-{SQL} over search queries}.
\newblock In \emph{Proceedings of the First Workshop on Interactive and
  Executable Semantic Parsing}, pages 34--45, Online. Association for
  Computational Linguistics.

\bibitem[{Shi et~al.(2020)Shi, Zhao, Boyd-Graber, Daum{\'e}~III, and
  Lee}]{shi-etal-2020-potential}
Tianze Shi, Chen Zhao, Jordan Boyd-Graber, Hal Daum{\'e}~III, and Lillian Lee.
  2020.
\newblock \href {https://doi.org/10.18653/v1/2020.findings-emnlp.167} {On the
  potential of lexico-logical alignments for semantic parsing to {SQL}
  queries}.
\newblock In \emph{Findings of the Association for Computational Linguistics:
  EMNLP 2020}, pages 1849--1864, Online. Association for Computational
  Linguistics.

\bibitem[{Sorokin and Gurevych(2018)}]{sorokin-gurevych-2018-mixing}
Daniil Sorokin and Iryna Gurevych. 2018.
\newblock \href {https://doi.org/10.18653/v1/S18-2007} {Mixing context
  granularities for improved entity linking on question answering data across
  entity categories}.
\newblock In \emph{Proceedings of the Seventh Joint Conference on Lexical and
  Computational Semantics}, pages 65--75, New Orleans, Louisiana. Association
  for Computational Linguistics.

\bibitem[{Tjong Kim~Sang and
  Veenstra(1999)}]{tjong-kim-sang-veenstra-1999-representing}
Erik~F. Tjong Kim~Sang and Jorn Veenstra. 1999.
\newblock \href {https://aclanthology.org/E99-1023} {Representing text chunks}.
\newblock In \emph{Ninth Conference of the {E}uropean Chapter of the
  Association for Computational Linguistics}, pages 173--179, Bergen, Norway.
  Association for Computational Linguistics.

\bibitem[{Wang et~al.(2020)Wang, Shin, Liu, Polozov, and
  Richardson}]{wang-etal-2020-rat}
Bailin Wang, Richard Shin, Xiaodong Liu, Oleksandr Polozov, and Matthew
  Richardson. 2020.
\newblock \href {https://doi.org/10.18653/v1/2020.acl-main.677} {{RAT-SQL}:
  Relation-aware schema encoding and linking for text-to-{SQL} parsers}.
\newblock In \emph{Proceedings of the 58th Annual Meeting of the Association
  for Computational Linguistics}, pages 7567--7578, Online. Association for
  Computational Linguistics.

\bibitem[{Wolf et~al.(2020)Wolf, Debut, Sanh, Chaumond, Delangue, Moi, Cistac,
  Rault, Louf, Funtowicz, Davison, Shleifer, von Platen, Ma, Jernite, Plu, Xu,
  Le~Scao, Gugger, Drame, Lhoest, and Rush}]{wolf-etal-2020-transformers}
Thomas Wolf, Lysandre Debut, Victor Sanh, Julien Chaumond, Clement Delangue,
  Anthony Moi, Pierric Cistac, Tim Rault, Remi Louf, Morgan Funtowicz, Joe
  Davison, Sam Shleifer, Patrick von Platen, Clara Ma, Yacine Jernite, Julien
  Plu, Canwen Xu, Teven Le~Scao, Sylvain Gugger, Mariama Drame, Quentin Lhoest,
  and Alexander Rush. 2020.
\newblock \href {https://doi.org/10.18653/v1/2020.emnlp-demos.6} {Transformers:
  State-of-the-art natural language processing}.
\newblock In \emph{Proceedings of the 2020 Conference on Empirical Methods in
  Natural Language Processing: System Demonstrations}, pages 38--45, Online.
  Association for Computational Linguistics.

\bibitem[{Wu et~al.(2020)Wu, Kao, Wu, Yin, and Liu}]{PERQ}
Zhiyong Wu, Ben Kao, Tien-Hsuan Wu, Pengcheng Yin, and Qun Liu. 2020.
\newblock \href {https://doi.org/10.1145/3336191.3371782} {\emph{PERQ:
  Predicting, Explaining, and Rectifying Failed Questions in KB-QA Systems}},
  page 663–671. Association for Computing Machinery, New York, NY, USA.

\bibitem[{Yang et~al.(2018)Yang, Liang, and Zhang}]{yang-etal-2018-design}
Jie Yang, Shuailong Liang, and Yue Zhang. 2018.
\newblock \href {https://aclanthology.org/C18-1327} {Design challenges and
  misconceptions in neural sequence labeling}.
\newblock In \emph{Proceedings of the 27th International Conference on
  Computational Linguistics}, pages 3879--3889, Santa Fe, New Mexico, USA.
  Association for Computational Linguistics.

\bibitem[{Yao et~al.(2019)Yao, Su, Sun, and Yih}]{yao-etal-2019-model}
Ziyu Yao, Yu~Su, Huan Sun, and Wen-tau Yih. 2019.
\newblock \href {https://doi.org/10.18653/v1/D19-1547} {Model-based interactive
  semantic parsing: A unified framework and a text-to-{SQL} case study}.
\newblock In \emph{Proceedings of the 2019 Conference on Empirical Methods in
  Natural Language Processing and the 9th International Joint Conference on
  Natural Language Processing (EMNLP-IJCNLP)}, pages 5447--5458, Hong Kong,
  China. Association for Computational Linguistics.

\bibitem[{Yu et~al.(2018)Yu, Zhang, Yang, Yasunaga, Wang, Li, Ma, Li, Yao,
  Roman, Zhang, and Radev}]{yu-etal-2018-spider}
Tao Yu, Rui Zhang, Kai Yang, Michihiro Yasunaga, Dongxu Wang, Zifan Li, James
  Ma, Irene Li, Qingning Yao, Shanelle Roman, Zilin Zhang, and Dragomir Radev.
  2018.
\newblock \href {https://doi.org/10.18653/v1/D18-1425} {{S}pider: A large-scale
  human-labeled dataset for complex and cross-domain semantic parsing and
  text-to-{SQL} task}.
\newblock In \emph{Proceedings of the 2018 Conference on Empirical Methods in
  Natural Language Processing}, pages 3911--3921, Brussels, Belgium.
  Association for Computational Linguistics.

\bibitem[{Zeng et~al.(2020)Zeng, Lin, Hoi, Socher, Xiong, Lyu, and
  King}]{zeng-etal-2020-photon}
Jichuan Zeng, Xi~Victoria Lin, Steven~C.H. Hoi, Richard Socher, Caiming Xiong,
  Michael Lyu, and Irwin King. 2020.
\newblock \href {https://doi.org/10.18653/v1/2020.acl-demos.24} {{P}hoton: A
  robust cross-domain text-to-{SQL} system}.
\newblock In \emph{Proceedings of the 58th Annual Meeting of the Association
  for Computational Linguistics: System Demonstrations}, pages 204--214,
  Online. Association for Computational Linguistics.

\bibitem[{Zhong et~al.(2017)Zhong, Xiong, and Socher}]{zhong2017seq2sql}
Victor Zhong, Caiming Xiong, and Richard Socher. 2017.
\newblock Seq2sql: Generating structured queries from natural language using
  reinforcement learning.
\newblock \emph{arXiv preprint arXiv:1709.00103}, 1:135--154.

\end{thebibliography}
\bibliographystyle{acl_natbib}

\clearpage

\appendix

\section{Implementation Details}
\label{sec:model-impl}

\subsection{Baselines Implementation}
We modified ETA since the vanilla version~\citep{liu-etal-2021-awakening} does not support ambiguous and unanswerable span detection.
For a fair comparison, we update the vanilla ETA in two ways.
Firstly, in the original inference part, the vanilla version of ETA applies the greedy linking algorithm to only keep the top 1 confidence score-related schema item (column or value). 
We change the selection process by allowing the top 3 candidates to be chosen, whose confidence score should be greater than the threshold.
We consider those spans with multi-grounding results as ambiguous spans.
Second, to enable ETA to handle unanswerable questions, we add a UNK column to the schema part and train the model to enable unanswerable span to get closer to the UNK column.
We consider those linked with the UNK column as unanswerable spans.
The heuristic-based baseline~\citep{sorokin-gurevych-2018-mixing} is n-gram matching via enumerating all n-gram ($n \leq 5$) phrases in a natural language question and links them to schema items by fuzzy string matching.
We consider a span as an ambiguous one when it can fuzzy match multiple results.
Similarly, if a noun phrase span can match no results, it is considered to unanswerable span.

\subsection{\ours Implementation}
Our \ours model consists of a BERT encoder, and three task modules, namely the concept prediction module, grounding module, and sequence labeling module.
We implement the first two modules following the implementation details mentioned in ~\citet{liu-etal-2021-awakening} and use the same hyperparameters.
In addition, the sequence labeling module is built by a dropout layer, a linear layer, and a CRF layer which is based on the open-source repository pytorch-crf\footnote{https://github.com/kmkurn/pytorch-crf}.
The response template for ambiguous questions is ``Oops, this question has multiple semantic meanings. {$X$} may refer to either "{concept1}", "{concept2}", or "{$\sColumn_{3}$}"''.
What's more, we design the template for unanswerable  questions as `` Sorry, we can’t find an answer for you since "$X$" cannot be mapped to any concepts in your table''.
Examples can be seen in \reffig{fig:intro-example}.

\subsection{Response Templates}
\label{sec:respnd-templ}
The response template for ambiguous questions is ``Oops, this question has multiple semantic meanings. {$X$} may refer to either "{concept1}", "{concept2}", or "{$\sColumn_{3}$}"''.
What's more, we design the template for unanswerable  questions as `` Sorry, we can’t find an answer for you since "$X$" cannot be mapped to any concepts in your table''.
Examples can be seen in \reffig{fig:intro-example}.

\subsection{Training Hyper-parameters}

For all experiments, we employ the AdamW optimizer and the default learning rate schedule strategy provided by Transformers library~\citep{wolf-etal-2020-transformers}.
The learning rate of other non-BERT layers is $1 \times 10^{-4}$.
The max training step is 100,000 and our training batch size is 35.
The training process last 6 hours on a single 16GB Tesla V100 GPU.

\section{Examples of \oursdataset dataset}
\label{sec:examples-dataset}
In this section, we demonstrate some good and bad cases of our \ours model prediction. 
Good  case examples are shown in \reftab{tab:amb-good-sp} and \reftab{tab:unk-good-sp}.
Bad case examples  are shown in \reftab{tab:amb-bad-sp} and \reftab{tab:unk-bad-sp}.

\begin{table}[htb]
    \centering
   \scalebox{0.5}{
        \begin{tabular}{c}
            \includegraphics[width=0.9\textwidth]{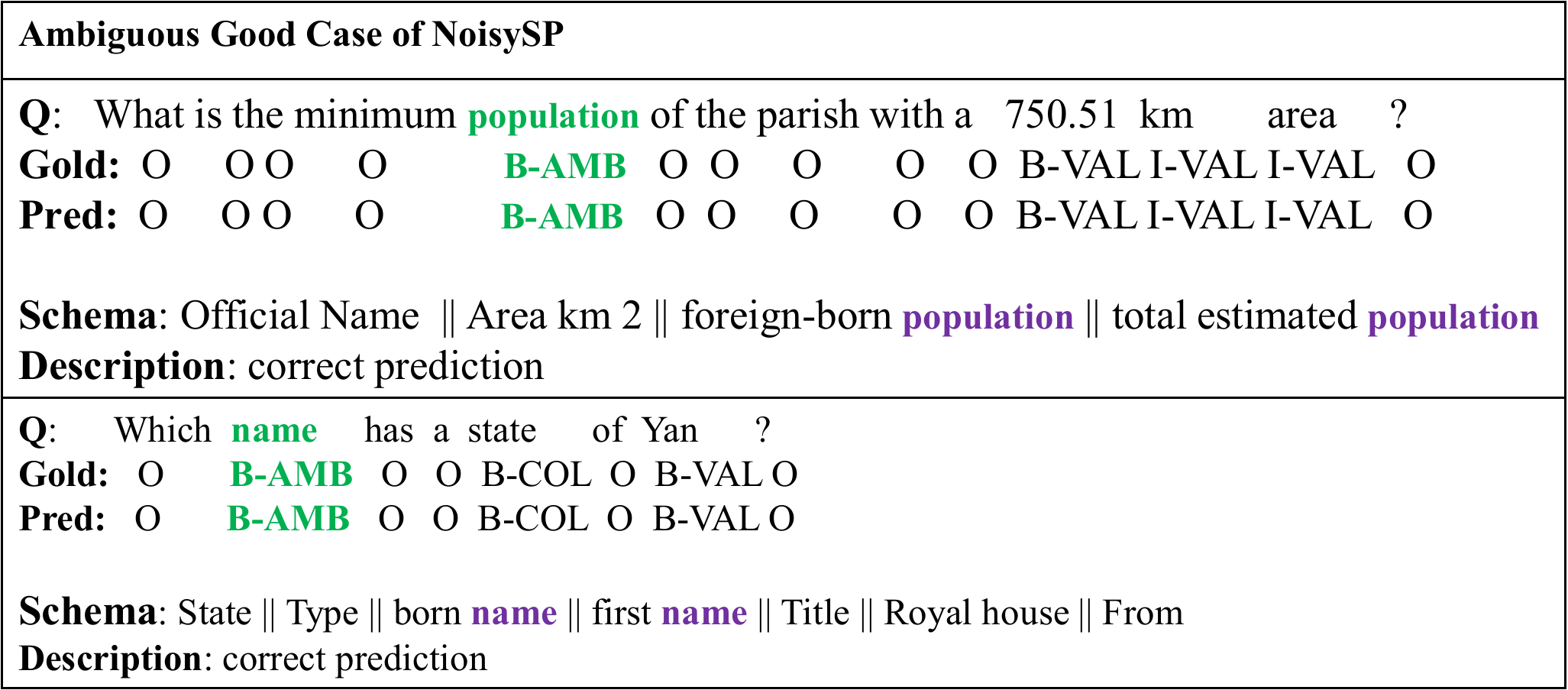}
        \end{tabular}
    }
	\caption{\AMBadj good cases by \ours on the \oursdataset  data.}
	\label{tab:amb-good-sp}
\end{table}

\begin{table}[htb]
    \centering
    \scalebox{0.5}{
        \begin{tabular}{c}
            \includegraphics[width=0.9\textwidth]{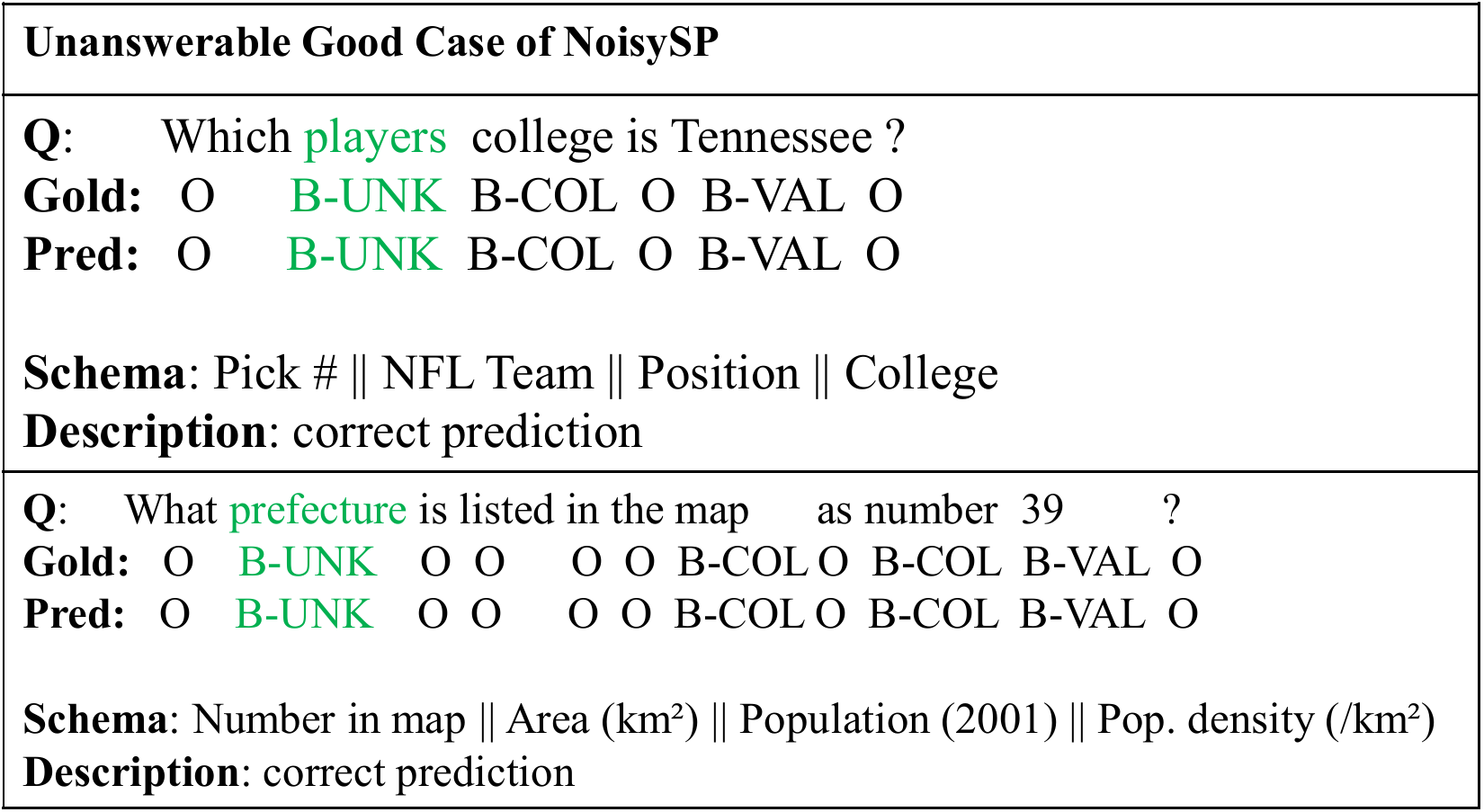}
        \end{tabular}
    }
	\caption{\Unk good cases by \ours on the \oursdataset  data.}
	\label{tab:unk-good-sp}
\end{table}

\clearpage

\begin{table}[h]
\small
    \centering
    \scalebox{0.5}{
        \begin{tabular}{c}
            \includegraphics[width=1.0\textwidth]{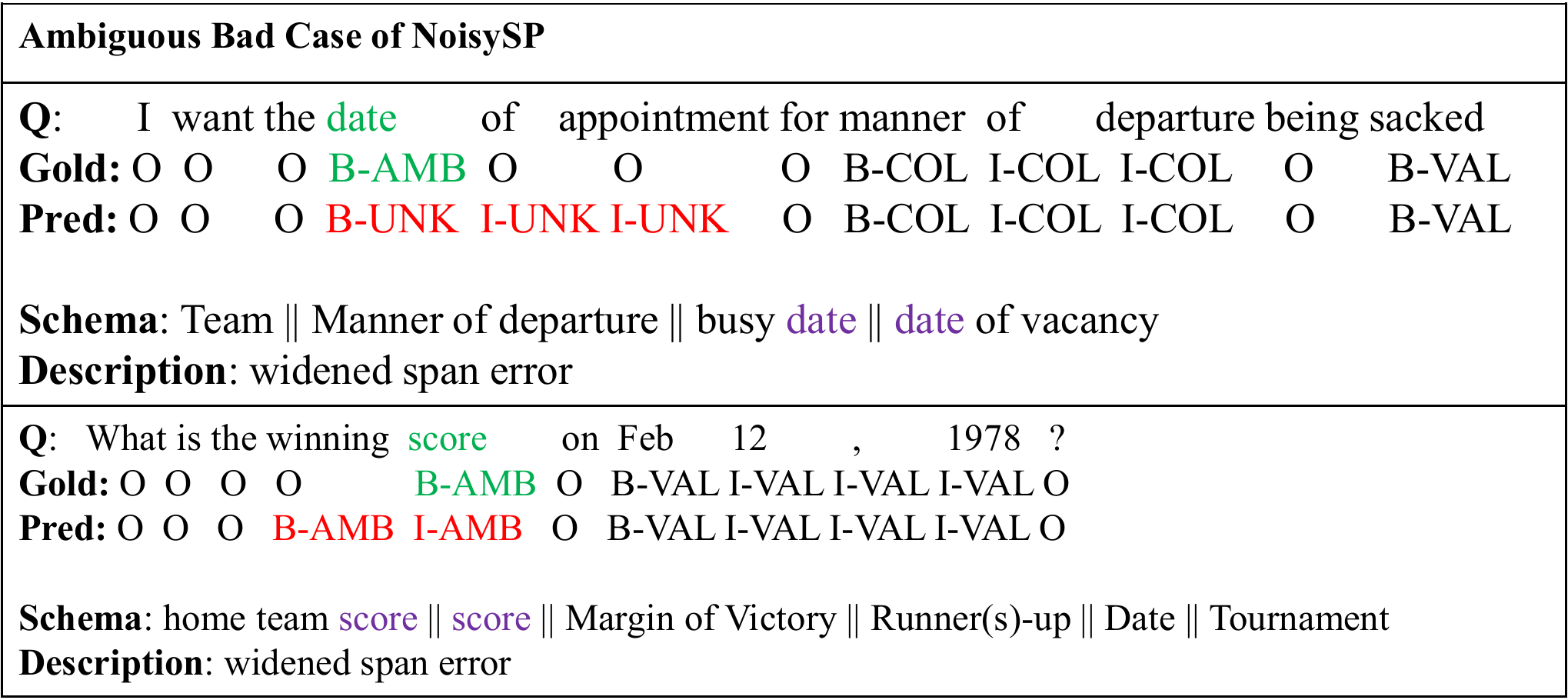}
        \end{tabular}
    }
	\caption{\AMBadj bad cases by \ours on the \oursdataset  data.}
	\label{tab:amb-bad-sp}
\end{table}

\begin{table}[h]
\small
    \centering
    \scalebox{0.5}{
        \begin{tabular}{c}
            \includegraphics[width=1.0\textwidth]{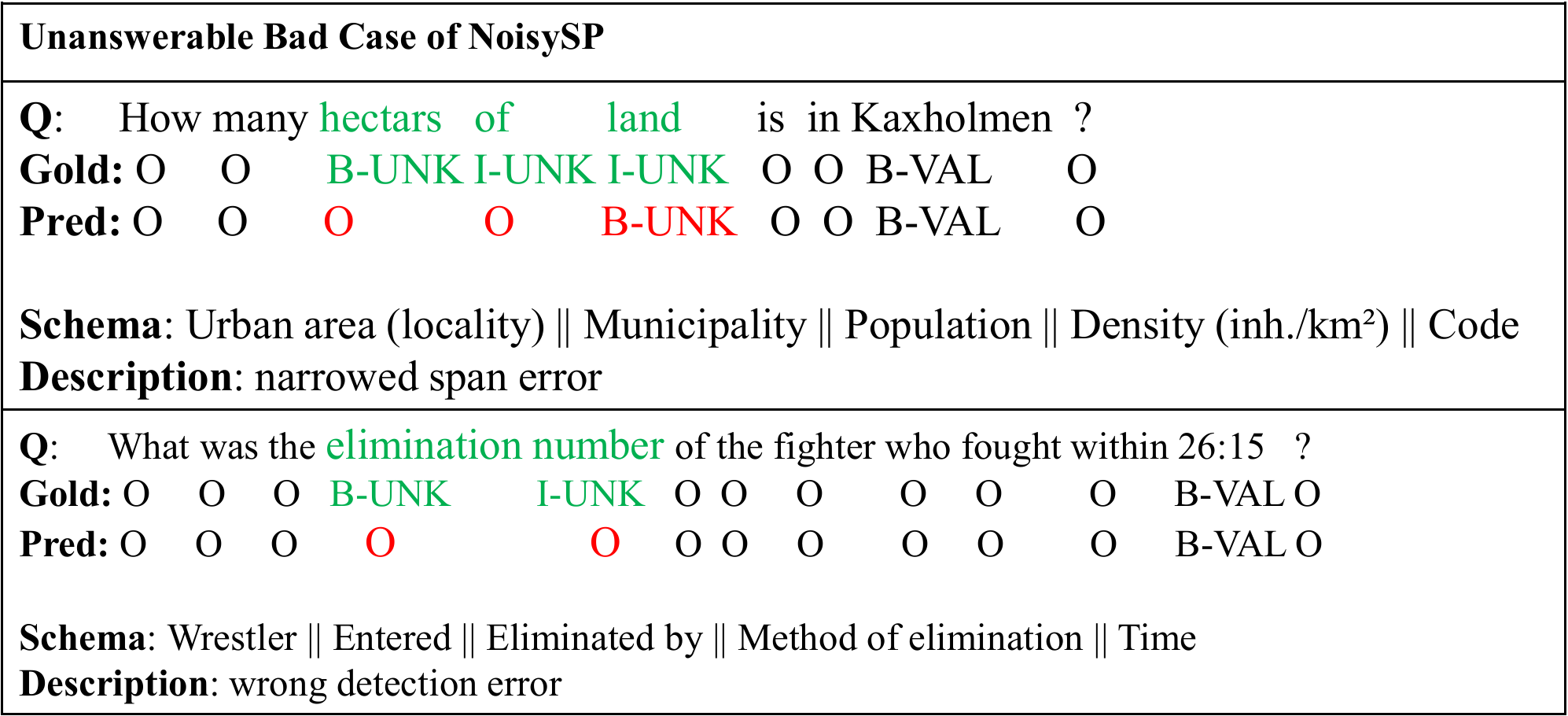}
        \end{tabular}
    }
	\caption{\Unk bad cases by \ours on the \oursdataset  data.}
	\label{tab:unk-bad-sp}
\end{table}

\section{Proposal for remaining 25\% problematic questions}
The remaining categories are (1) Value Ambiguity (10\%); (2) Value Unanswerable (7\%); (3) Calculation Unanswerable (6\%); and (4) Out of Scope (2\%). Although our data generation method does not cover these categories, our DTE model trained with our generated dataset NoisySP can generalize to questions of these categories (results are shown in \refsec{sec:generaization-to-realistic-data}), especially for Value Ambiguity and Value Unanswerable. This is because columns and values are treated as concepts that share the same pattern.

For the Calculation Unanswerable category, formulaic knowledge is needed to inject the model with the necessary background information. Existing works, such as KnowSQL~\cite{dou-etal-2022-towards}, are intended to solve this kind of problem.

As for the Out of Scope problem, which usually calls for graphic operation or other unsupported operations, it can be easily handled with a blacklist or a simple classifier.

\end{document}